\documentclass{article}
\usepackage{ijcai19}

\usepackage{times}
\usepackage{soul}
\usepackage{url}
\usepackage[hidelinks]{hyperref}
\usepackage[utf8]{inputenc}
\usepackage[small]{caption}
\usepackage{graphicx}
\usepackage{amsmath}
\usepackage{booktabs}
\usepackage{algorithm}
\usepackage{algorithmic}
\usepackage{subfigure}
\usepackage{multirow}
\title{Hierarchical Attention Generative Adversarial Networks for Cross-domain Sentiment Classification}

\author{
Yuebing Zhang$^{1,2}$\and
Duoqian Miao$^{1,2}$\footnote{Contact Author}\and
Jiaqi Wang$^{1,2}$\\
\affiliations
$^1$Department of Computer Science and Technology, Tongji University, Shanghai, China\\
$^2$Key Laboratory of Embedded System and Service Computing, Ministry of Education, Shanghai, China\\
\emails
yuebing\_zhang@hotmail.com,
dqmiao@tongji.edu.cn,
queeny\_wang@163.com
}

\begin{document}

\maketitle

\begin{abstract}
  Cross-domain sentiment classification (CDSC) is an importance task in domain adaptation and sentiment classification. Due to the domain discrepancy, a sentiment classifier trained on source domain data may not works well on target domain data. In recent years, many researchers have used deep neural network models for cross-domain sentiment classification task, many of which use Gradient Reversal Layer (GRL) to design an adversarial network structure to train a domain-shared sentiment classifier. Different from those methods, we proposed Hierarchical Attention Generative Adversarial Networks (HAGAN) which alternately trains a generator and a discriminator in order to produce a document representation which is sentiment-distinguishable but domain-indistinguishable. Besides, the HAGAN model applies Bidirectional Gated Recurrent Unit (Bi-GRU) to encode the contextual information of a word and a sentence into the document representation. In addition, the HAGAN model use hierarchical attention mechanism to optimize the document representation and automatically capture the pivots and non-pivots. The experiments on Amazon review dataset show the effectiveness of HAGAN.
\end{abstract}

\section{Introduction}

 Sentiment classification is an important task in natural language processing (NLP), and it aims at identifying the overall sentiment polarity of a subjective text \cite{Pang2008Opinion}. Researchers applied machine learning methods with handcrafted features for sentiment classification \cite{wang2012baselines} before the popularity of deep learning. Recently, many researchers applied neural network models for sentiment classification, and have achieved good classification performance in the single domain with labeled data \cite{socher2013recursive}. However, neural networks, which are supervised learning methods, can't work well in the case where the labels of data are time-consuming or expensive. Therefore, CDSC, in which target domain data has no or few labels so we can only train a classifier in labeled source domain data and adapt it to target domain, has become a hot research direction. However, the expression of users' emotion varies widely across different domains. Sentiment words in one domain may not work in another, even not appear. On the other hand, a sentiment word may expresses different emotions in different domains. Due to the domain discrepancy, the sentiment classifier trained on the source domain data may not work well if it's directly applied to the target domain.

 The key point of CDSC is how to use the labeled date in source domain and the unlabeled data in target domain to train a domain-independent sentiment classifier. In the early days of CDSC research, the researchers proposed \emph{pivot} and \emph{non-pivot}, which denote the sentiment word that works in both source and target domains and the sentiment word that works in only one domain, respectively. In the era of discrete text feature, many works build the bridge between source and target domains by using the pivots, and two representative works are Structural Correspondence Learning (SCL) \cite{Blitzer2007Biographies} and Spectral Feature Alignment (SFA) \cite{Pan2010Cross}. However, SCL and SFA need to manually select the pivots for each source-target domain pair which is time-consuming. Besides, those methods use discrete text feature and linear classifier, which results in poor classification performance. With the development of deep learning, many neural network models were proposed for CDSC. The key to the neural network models for CDSC is to generate a domain-shared feature representation so that the sentiment classifier works well in both source and target domains. Many researchers apply the unlabeled data for training to learn a shared feature representation, e.g., Stacked Denoising Autoencoders (SDA) \cite{glorot2011domain} and Neural Networks with Auxiliary Task (AuxNN) \cite{yu2016learning}. On the other hand, many adversarial methods were proposed for CDSC, e.g., Domain-Adversarial training of Neural Networks (DANN) \cite{ganin2015unsupervised,ganin2016domain}, Adversarial Memory networks (AMN)\cite{Li2017End} and Hierarchical Attention Transfer Networks (HATN) \cite{li2018hierarchical}. Different from those adversarial methods, the proposed method HAGAN apply Generative Adversarial Networks (GAN) \cite{goodfellow2014generative} to alternately train a generator and a discriminator, instead of using GRL \cite{ganin2015unsupervised,ganin2016domain}. GAN is widely used for image generation in Computer Vision (CV), but it is hard to be directly applied to text generation in NLP, because text features are discrete and GAN can not calculate the loss function values of an incomplete sequence. However, we can apply GAN for encoded real-value feature space alignment in CDSC. The proposed HAGAN model consists of a generator and a discriminator, which are alternately trained. The discriminator has two objectives: sentiment classification and domain classification. The generator attempts to fool the discriminator on the domain classification subtask, while generating a sentiment-distinguishable document representation. In the generator, Bi-GRU is used to encode the contextual information of a word and a sentence into the document representation \cite{yang2016hierarchical}, and hierarchical attention mechanism is applied to optimize the document representation and automatically capture the pivots and non-pivots. The experiments on Amazon review dataset show that the HAGAN model can reduce the difference of document representations in different domains, and has a good performance in classification accuracy.

\section{Related Work}

 There are some traditional methods for CDSC. Blitzer et al. \cite{Blitzer2007Biographies} proposed SCL, which uses pivot prediction task to learn a shared feature representation for source and target domains. Pan et al. \cite{Pan2010Cross} proposed SFA to construct the alignment between the pivots and non-pivots by using the cooccurrence between them, in order to build a bridge between source and target domains. In general, these traditional methods need to manually select the pivots which means the performance of these methods depends on the choice of pivot. Besides, manually selecting pivots for each source-target domain pair is time-consuming.

 With the development of deep learning, many neural network methods were proposed for CDSC. Glorot et al. \cite{glorot2011domain} proposed SDA to learn a shared feature representation for all domains by using a large amount of unlabeled documents from many domains. Yu et al. \cite{yu2016learning} used the data in both source and target domains to construct auxiliary prediction tasks which are highly correlated with main task. However, these two methods can not identify the pivots. Recently, there are some adversarial methods proposed for CDSC. Ganin et al. \cite{ganin2015unsupervised,ganin2016domain} proposed DANN which use a GRL to reverse the gradient direction. The GRL can help the neural networks to produce domain-confused representations so that the sentiment classifier works well on both domains. The GRL is the key point of the existing adversarial training methods for CDSC. Some variants of GRL-based adversarial methods were subsequently proposed. Li et al. \cite{Li2017End} applied attention mechanism in word encoding in order to directly identify the pivots. Furthermore, Li et al. \cite{li2018hierarchical} proposed HATN, which is a two-stage neural network model, to automatically identify the non-pivots. In general, the existing adversarial methods for CDSC use GRL to adversarially train a neural network. Different from those methods, the proposed HAGAN model in this work applies GAN architecture to alternately train a generator and a discriminator, in order to generate a document representation which is sentiment-distinguishable but domain-indistinguishable.

\section{Method}
 In this section, we introduce the proposed HAGAN model. We first present the problem definition and notations, after that we present the overview of the model, and then we detail the model with all components.

\subsection{Problem Definition and Notations}
We are given two domains $D_s$ and $D_t$ which denote a source domain and a target domain, respectively. In $D_s$, we have a set of labeled data $X_s^l = \{x_s^i\}_{i = 1}^{N_s^l}$ and $\{\hat{y}_s^i \}_{i=1}^{N_s^l}$ as well as a set of unlabeled data $X_s^u = \{x_s^i\}_{i = N_s^l + 1}^{N_s}$, where $X_s = X_s^l \cup X_s^u$. In $D_t$, we have a set of unlabeled data $X_t = \{ x_t^j \}_{j = 1}^{N_t}$. The goal of cross-domain sentiment classification is to train a sentiment classifier on $X_s^l$ and using $X_s^u$ and $X_t$ to adopt the classifier to predict the sentiment polarity of $X_t$.

\subsection{An Overview of HAGAN}
\begin{figure*}
  \centering
  \includegraphics[width=500px]{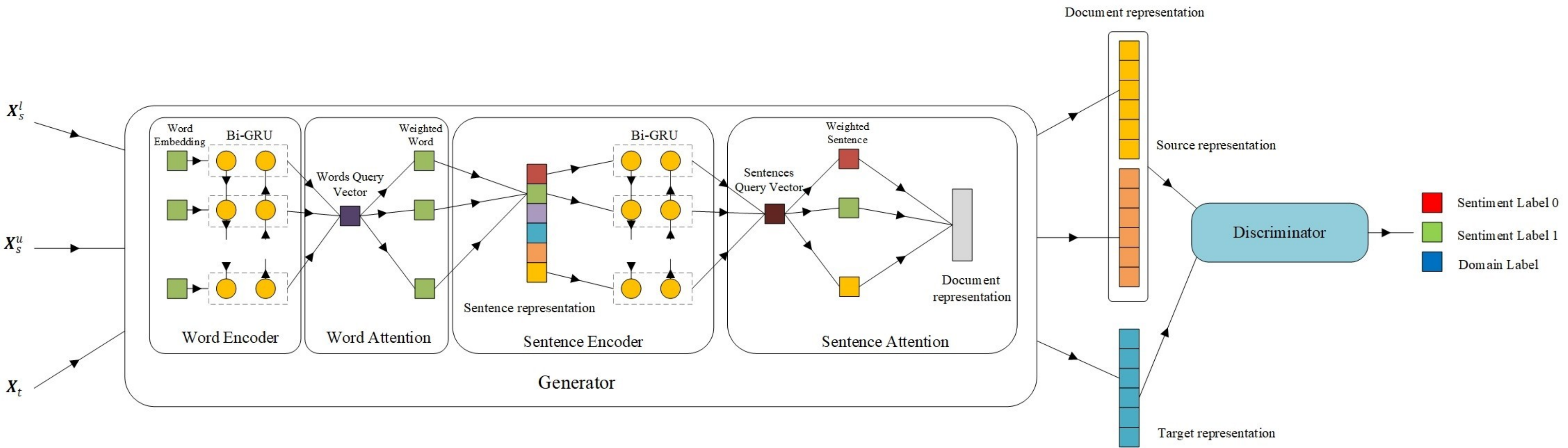}\\
  \caption{The framework of the HAGAN model.}\label{fig1}
\end{figure*}
The HAGAN model uses GAN to generate a document representation which is sentiment-distinguishable but domain-indistinguishable. Besides, the HAGAN model applies hierarchical attention mechanism for document representation generation to ensure interpretability of document representation. The attention mechanism can help to automatically identify the pivots and non-pivots in the neural network models, and the hierarchical attention mirrors the hierarchical structure of text, e.g., word-sentence-document granular structure used in this work. As shown in Fig.~\ref{fig1}, all the samples, including $X_s^l$, $X_s^u$ and $X_t$, are entered into the generator. The generator consists of four components: word encoder, word attention, sentence encoder and sentence attention. The generator outputs source representations and target representations which denote the document representations of the samples from source and target domains, respectively. The discriminator, which is a Multi-Layer Perception (MLP), has two objectives, one is to classify $X_s^l$ into several categories according to the sentiment labels (positive and negative for binary classification), the other is to classify $X_s \cup X_t$ into two categories, i.e., source or target. The HAGAN model is different from the original GAN. The HAGAN model uses real text samples as the input of generator, while the original GAN use random noise as the input to generate fake images. The discriminator of the HAGAN model has $N+1$ outputs for $N$ sentiment classification in order to process two subtasks which are domain-classification and sentiment-classification, while the discriminator of original GAN has only two outputs which denote real and fake, respectively.

\subsection{Components}

In this section, we detail each component of the HAGAN model.

\subsubsection{Word Encoder}

Suppose that a document $D$ is made up of $L$ sentences, each sentence is make up of $T_i$ words, where $i \in [1, L]$. Given the words $w_{ij}$ in sentence $i$, $j \in [1, T_i]$, we first map each word into its embedding vector trough an embedding matrix $M_e$, $e_{ij} = M_e w_{ij}$. We use a Bi-GRU to get the representation of words by summarizing information from two directions of the word sequence, so that the word representations incorporate contextual information of current word. The Bi-GRU contains the forward GRU $\overrightarrow{f}$ which reads the sentence $i$ from $w_{i1}$ to $w_{iT_i}$ and the backward GRU $\overleftarrow{f}$ which reads the sentence $i$ from $w_{iT_i}$ to $w_{i1}$.
\begin{align}
  e_{ij} = M_e w_{ij}, j \in [1, T_i] \label{eq1} \\
  \overrightarrow{h_{ij}} = \overrightarrow{f}(e_{ij}), j \in [1, T_i] \label{eq2} \\
  \overleftarrow{h_{ij}} = \overleftarrow{f}(e_{ij}), j \in [T_i, 1] \label{eq3}
\end{align}
We concatenate the forward hidden state $\overrightarrow{h_{ij}}$ and the backward hidden state $\overleftarrow{h_{ij}}$ to get a bidirectional hidden state representation of the given word $w_{ij}$, i.e., $h_{ij} = [\overrightarrow{h_{ij}}, \overleftarrow{h_{ij}}]$, in which the contextual information of current word has been encoded.

\subsubsection{Word Attention}

In sentiment classification task, each word contributes differently to the sentence representation. Thus we apply attention mechanism in word level to calculate the importance of each word in current sentence for sentiment classification, and then integrate the weighted word representations to form a sentence vector.
\begin{align}
  \alpha_{ij} &= \frac{\exp (h_{ij}^\top q_w)}{\sum_{j} \exp(h_{ij}^\top q_w)} \label{eq4}\\
  s_i &= \sum_j \alpha_{ij}h_{ij} \label{eq5}
\end{align}
The importance weight of a word $w_{ij}$ is calculated by the similarity of its bidirectional hidden state representation $h_{ij}$ and the word-level query vector $q_w$. Then we compute the sentence vector $s_i$ as the weighted sum of $h_{ij}$ according to the importance weight $\alpha_{ij}$. The word-level query vector $q_w$ can seen as a high-level representation of a fixed query ``What is the important word in this sentence for sentiment classification'', and it is randomly initialized and jointly learned during the training process.

\subsubsection{Sentence Encoder}

Given the sentence vectors $s_i$, we can calculate the document representation in a similar way. We also use a Bi-GRU to encode the sentences:
\begin{align}
  \overrightarrow{h_{i}} = \overrightarrow{f}(s_{i}), i \in [1, L] \label{eq6}\\
  \overleftarrow{h_{i}} = \overleftarrow{f}(s_{i}), i \in [L, 1] \label{eq7}
\end{align}
We concatenate $\overrightarrow{h_{i}}$ and $\overleftarrow{h_{i}}$ to get the bidirectional hidden state representation of sentence $i$, i.e., $h_i = [\overrightarrow{h_{i}}, \overleftarrow{h_{i}}]$, in which the contextual information of current sentence has been encoded.

\subsubsection{Sentence Attention}

Similar with word attention mechanism, each sentence contributes differently to form a document representation for sentiment classification. Thus we again apply the attention mechanism in sentence level to measure the importance of each sentence for sentiment classification task. The document representation $d$ is computed as the weighted sum of $h_i$ according to the importance weight $\alpha_i$.
\begin{align}
  \alpha_{i} &= \frac{\exp (h_{i}^\top q_s)}{\sum_{i} \exp(h_{i}^\top q_s)} \label{eq8}\\
  d &= \sum_i \alpha_{i}h_{i} \label{eq9}
\end{align}
The sentence query vector $q_s$ is similar with the word query vector $q_w$, and it is initialized randomly and jointly learned during the training process.

\subsubsection{Discriminator}

The discriminator in the HAGAN model is a MLP with a softmax layer as the output. For simplicity, we describe how discriminator works in binary sentiment classification case (positive or negative).
\begin{equation}\label{eq10}
  p = \mathrm{softmax}(\mathrm{tanh}(W_D d + b_D))
\end{equation}
In binary sentiment classification, the discriminator output $p$ consists of three neuron outputs, $p_p$, $p_n$, $p_t$, which respectively denote positive sample, negative sample and target domain sample. For the labeled data in $X_s^l$ which is for sentiment classification subtask, we focus on $p^{sen} = \mathrm{softmax}([p_p, p_n])$. For the unlabeled data $X_s^u$ and $X_t$ which is for domain classification subtask, we focus on $p^{dom} = [p_p + p_n, p_t]$ in which $p_p + p_n$ can be seen as the probability that the current sample is from source domain.
\subsection{Training of HAGAN}

The training strategy of the HAGAN model is similar with the original GAN, that is generator-discriminator alternation training. We first freeze the parameters of the generator and train the discriminator, then freeze the parameters of the discriminator and train the generator, and alternate the two steps until the Nash equilibrium is reached. The loss used to train the discriminator is defined as:
\begin{equation}\label{eq11}
  L_D = L_{sen} + \lambda_D L_{dom}
\end{equation}
where $\lambda_D$ is to balance the sentiment loss and domain loss. The sentiment loss $L_{sen}$ is to minimize the cross-entropy for the labeled data $X_s^l$ in source domain:
\begin{equation}\label{eq12}
  L_{sen} = - \frac{1}{N_s^l} \sum_{i=1}^{N_s^l}y_i\mathrm{ln}\hat{y}_i+(1-y_i)\mathrm{ln}(1-\hat{y}_i)
\end{equation}
where $y_i = \mathrm{argmax} (p_i^{sen}), \hat{y}_i \in \{0, 1\}$ are the sentiment prediction and golden sentiment label of the $i$th source labeled sample, respectively. Similarly, the domain loss $L_{dom}$ is to minimize the cross-entropy for the unlabeled data $X_s \cup X_t$ (labeled and unlabeled data are both used in this step) in source and target domains:
\begin{equation}\label{eq13}
  L_{dom} = - \frac{1}{N_s+N_t} \sum_{i=1}^{N_s+N_t}d_i\mathrm{ln}\hat{d}_i+(1-d_i)\mathrm{ln}(1-\hat{d}_i)
\end{equation}
where $d_i = \mathrm{argmax} (p_i^{dom}), \hat{d}_i \in \{0, 1\}$ are the domain prediction and golden domain label of the $i$th sample, respectively.

The loss used to train the generator is defined as:
\begin{equation}\label{eq14}
  L_G = L_{sen} + \lambda_G^1 L_{dom}^c + \lambda_G^2 L_{ent}
\end{equation}
where $\lambda_G^1$ and $\lambda_G^2$ are to balance the sentiment loss, domain confusion loss and entropy loss. The domain confusion loss $L_{dom}^c$ is to minimize the cross-entropy for the unlabeled data $X_t$ with masked domain labels in target domain:
\begin{equation}\label{eq15}
  L_{dom}^c = - \frac{1}{N_t} \sum_{i=1}^{N_t}d_i\mathrm{ln}\tilde{d}_i+(1-d_i)\mathrm{ln}(1-\tilde{d}_i)
\end{equation}
where $\tilde{d}_i$ is the masked domain label. In this step, all the domain labels of the $X_t$ are masked as ``source'', so that the generator can generate a domain-shared document representation which attempts to confuse the discriminator.

The entropy loss is to minimize the entropy of the sentiment prediction distribution of the unlabeled target domain data $X_t$:
\begin{equation}\label{eq16}
  L_{ent} = - \frac{1}{N_t} \sum_{i=1}^{N_t} \sum_{j=1}^{C} p_{ij}^{sen} \mathrm{ln}p_{ij}^{sen}
\end{equation}
where $C$ is the number of sentiment labels, $C=2$ for binary sentiment classification. The entropy loss can help to maximize the margins between the target domain data and the decision boundaries, and increase the prediction confidence of the target domain data.
\section{Experiment}

\subsection{Experimental Settings}

We conduct the experiments on the Amazon review dataset, which has been widely used for cross-domain sentiment classification. This dataset contains four different domains: Book (B), DVD (D), Electronics (E), and Kitchen (K). We consider the binary sentiment classification task to predict whether a review is positive (higher than 3 stars) or negative (3 stars and lower than 3 stars). Each domain consists of 1000 positive reviews and 1000 negative reviews. We allow 4000 unlabeled reviews to be used for both source and target domains. We construct 12 cross-domain sentiment classification tasks and split the labeled data in each domain into a training set of 1600 reviews (800 positive and 800 negative) and a test set of 400 reviews (200 positive and 200 negative) for sentiment classification subtask. All the labeled and unlabeled data are used for training in domain classification subtask.

\subsection{Implementation Details}

We experimented with the pre-trained BERT\footnote{https://github.com/google-research/bert} word embeddings, which are 768-dimensional. In BERT model, a word maps different embeddings in different sentences, thus we need to use BERT to generate each word embedding in the reviews before we input them into the HAGAN model, instead of applying a embedding layer to do that. For each pair of domains, the vocabulary consists of the top 10000 most frequent words. We use NLTK to split the reviews into sentences. The dimensions of sentence representations and document representations are set to 1000 and 2000, respectively. The GRU contains 100 units, and uses \emph{tanh} as the activation function. $\lambda_D$ is set to $1$. $\lambda_G^1$ and $\lambda_G^2$ are set to $0.2$ and $0.02$, respectively. The discriminator has two hidden layer whose width are 512 and 256, respectively, and \emph{tanh} activation function and dropout of $0.25$ are used here. We trained the generator and discriminator on batch with the batch size of 100. RMSProp optimizer with 0.0005 learning rate is used for both generator and disciminator training.

\subsection{Performance Comparison}

We compare the HAGAN model with the following baselines:
\begin{itemize}
  \item \textbf{Naive} is a non-domain-adaptive baseline with bag-of-words representations and SVM classifier trained on the labeled data from source domain.
  \item \textbf{mSDA} \cite{chen2012marginalized} is one of the state-of-the-art domain adaptation method based on discrete input features. Top 1000 bag-of-words features are kept as pivot features.
  \item \textbf{NaiveNN} is a non-domain-adaptive CNN model \cite{Kim2014Convolutional} trained on the labeled data from source domain.
  \item \textbf{AuxNN} \cite{yu2016learning} constructs two auxiliary prediction tasks to help CNN encoder to generate a domain-share representation.
  \item \textbf{ADAN} \cite{chen2018adversarial} expoits adversarial training to transfer the knowledge of resource-rich source language to low-resource language. In this work, we adapt it to cross-domain sentiment classification tasks by apply a domain classifier instead of language classifier. We apply a CNN as the encoder.
  \item \textbf{AMN} \cite{Li2017End} applies attention mechanism to generate the review representations, and exploit GRL to adversarially train a sentiment classifier and a domain classifier.
\end{itemize}
In addition to the above baselines, we also show results of a variant \textbf{HAGAN-C} of our model \textbf{HAGAN}. \textbf{HAGAN-C} applies a CNN \cite{Kim2014Convolutional} as discriminator instead of a MLP.

\begin{table*}
\centering
\begin{tabular}{c|cccccc|cc}
  \hline
  Tasks & Naive & mSDA & NaiveNN & AuxNN & ADAN & AMN & HAGAN & HAGAN-C \\
  \hline
  D$\to$B & 75.20 & 78.50 & 81.12 & 80.80 & 81.70 & 81.52 & 81.22 & \textbf{81.69} \\
  E$\to$B & 68.85 & 76.15 & 77.75 & 78.00 & 78.55 & 77.80 & 79.05 & \textbf{79.23} \\
  K$\to$B & 70.00 & 75.65 & 78.37 & 77.85 & 79.25 & \textbf{79.37} & 78.52 & 78.99 \\
  \hline
  B$\to$D & 77.15 & 80.60 & 80.98 & 81.75 & 82.30 & 81.32 & 82.07 & \textbf{82.38} \\
  E$\to$D & 69.50 & 76.30 & 77.12 & 80.65 & 79.70 & 77.51 & \textbf{81.00} & 80.65 \\
  K$\to$D & 71.40 & 76.05 & 79.35 & 78.90 & 80.45 & 80.03 & 80.83 & \textbf{80.91} \\
  \hline
  B$\to$E & 72.15 & 75.55 & 77.68 & 76.40 & 77.60 & 80.07 & 79.87 & \textbf{80.12} \\
  D$\to$E & 71.65 & 76.00 & 78.32 & 77.55 & 79.70 & 80.00 & 80.57 & \textbf{80.99} \\
  K$\to$E & 79.75 & 84.20 & 84.98 & 84.05 & \textbf{86.85} & 81.97 & 85.94 & 85.23 \\
  \hline
  B$\to$K & 73.50 & 75.95 & 77.10 & 78.10 & 76.10 & 81.00 & 81.25 & \textbf{82.00} \\
  D$\to$K & 72.00 & 76.30 & 78.79 & 80.05 & 77.35 & \textbf{83.88} & 81.73 & 81.50 \\
  E$\to$K & 82.80 & 84.45 & 85.06 & 84.15 & 83.95 & \textbf{87.10} & 84.30 & 84.99 \\
  \hline
  Avg & 73.66 & 77.98 & 79.72 & 79.85 & 80.29 & 80.96 & 81.36 & \textbf{81.56} \\
  \hline
\end{tabular}
\caption{Accuracies on the Amazon dataset.}
\label{tab1}
\end{table*}
Table~\ref{tab1} reports the accuracies of baselines and the proposed methods on Amazon dataset. The proposed methods outperform all the baselines on average in terms of accuracy. Naive and mSDA perform poorly because of discrete input features. NaiveNN which does not use any target domain data performs good on average, which is due to the fact that BERT embeddings contain rich context information extracted from a large amount of corpus. AuxNN uses target domain data to construct auxiliary tasks, and performs a little better than NaiveNN which use no target domain data. The proposed methods HAGAN and HAGAN-C perform slightly better than AMN. Same as adversarial methods, HAGAN applies GAN structure to implement adversarial training, while AMN use GRL. Besides, HAGAN uses a hierarchical attention mechanism which maps to word-sentence-document structure of text, while AMN use 3 layer memory networks, which ignores the inner structure of text. And additionally HAGAN applies Bi-GRU to extract context information when constructing high-level representations. The adversarial training method of ADAN is similar with that of HAGAN. But HAGAN unifies the adversarial training into GAN, and use the training techniques of GAN to improve the performance. Besides, the encoder of ADAN is too simple. HAGAN-C outperforms HAGAN on average. The reason is the discriminating ability of CNN is better than MLP. We need to balance the ability of the generator and discriminator. The HAGAN model performs poorly if the discriminator is much weaker than the generator.

\subsection{Visualization of Representation}

\begin{figure*}
\centering

\subfigure[Naive HAN model.]{
\begin{minipage}{0.45\linewidth}
\centering
\includegraphics[width=240px]{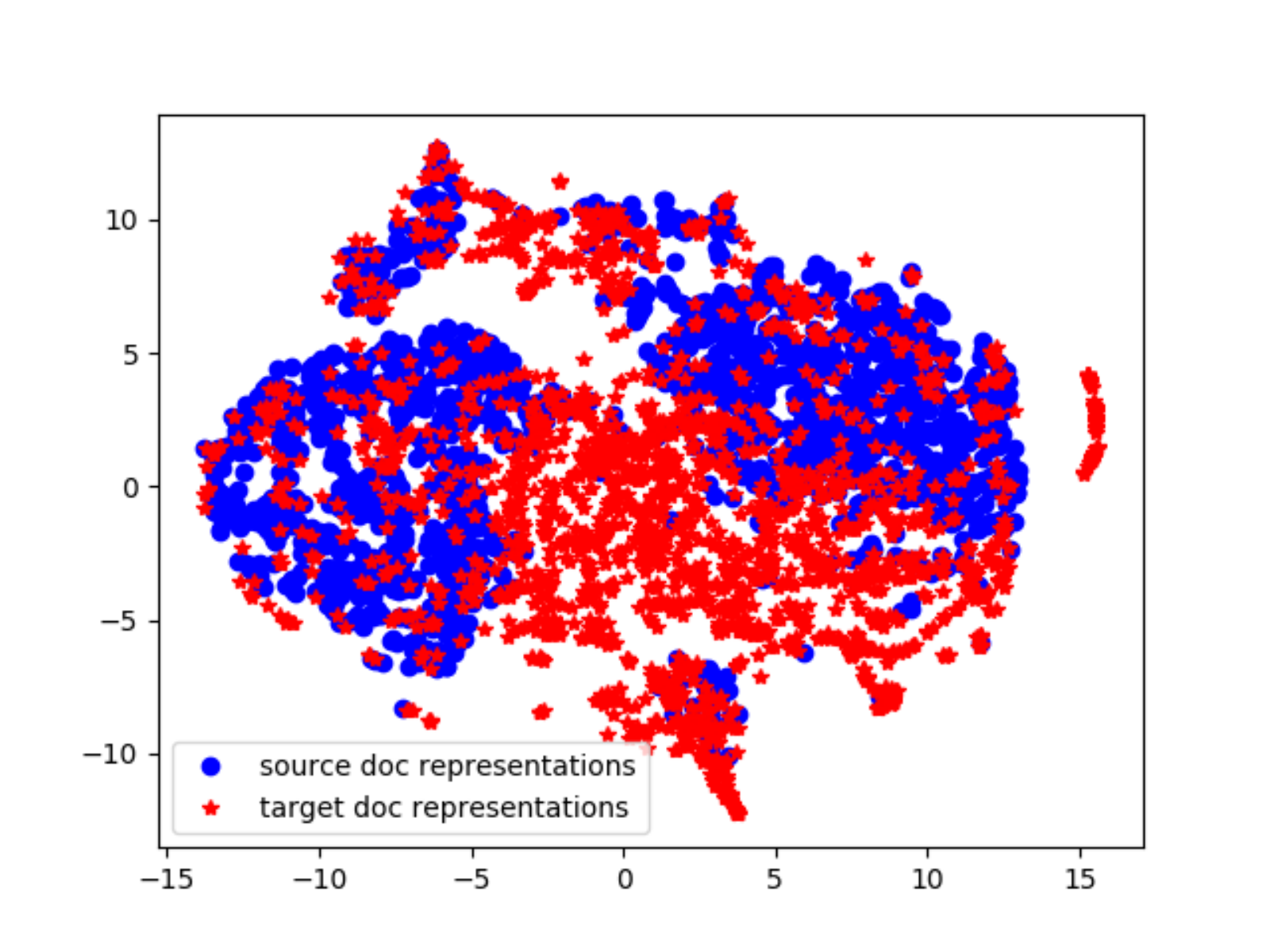}
\end{minipage}%
}%
\subfigure[HAGAN model.]{
\begin{minipage}{0.45\linewidth}
\centering
\includegraphics[width=240px]{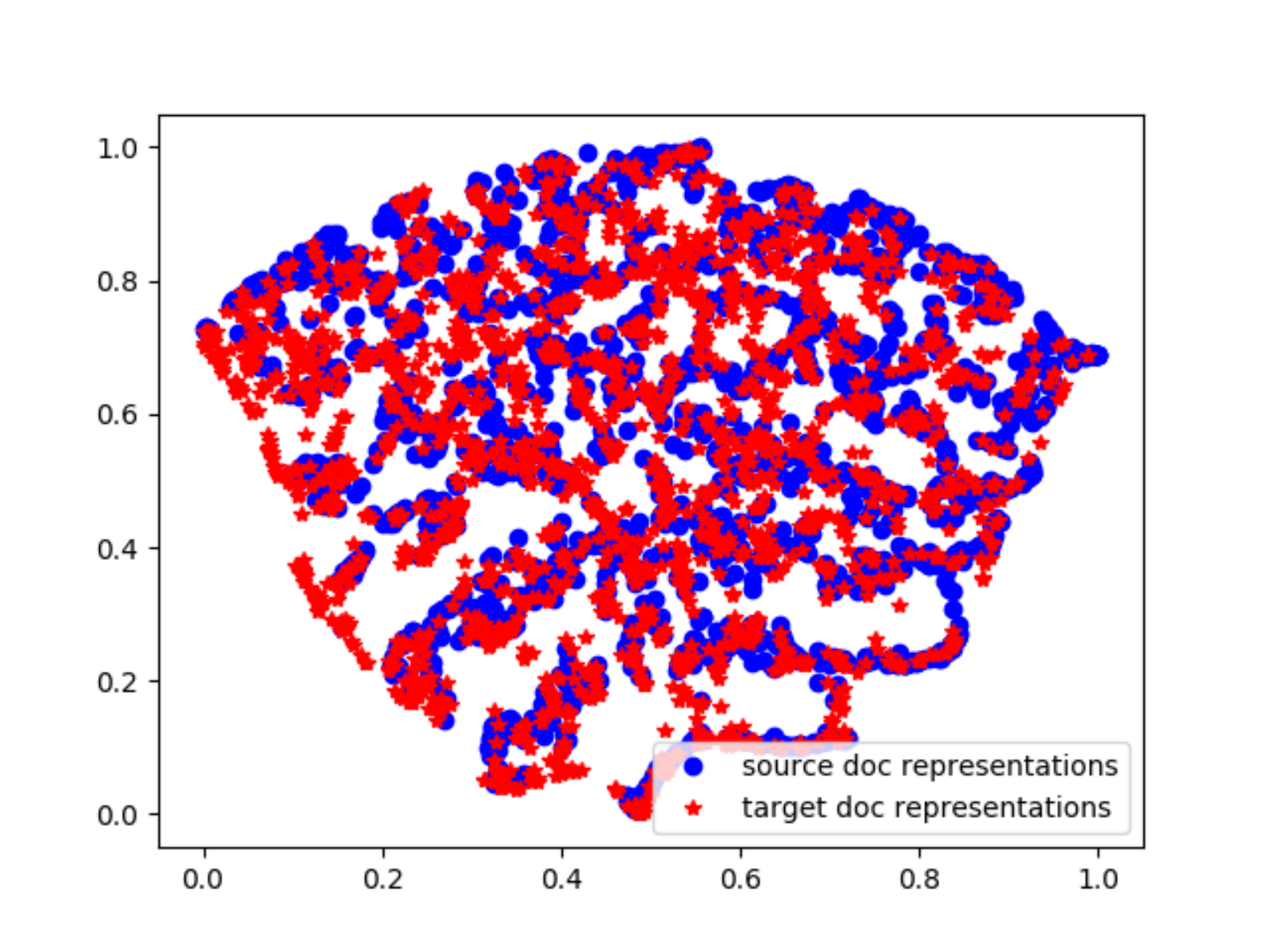}
\end{minipage}%
}%
\quad
\subfigure[Source domain in HAGAN.]{
\begin{minipage}{0.45\linewidth}
\centering
\includegraphics[width=240px]{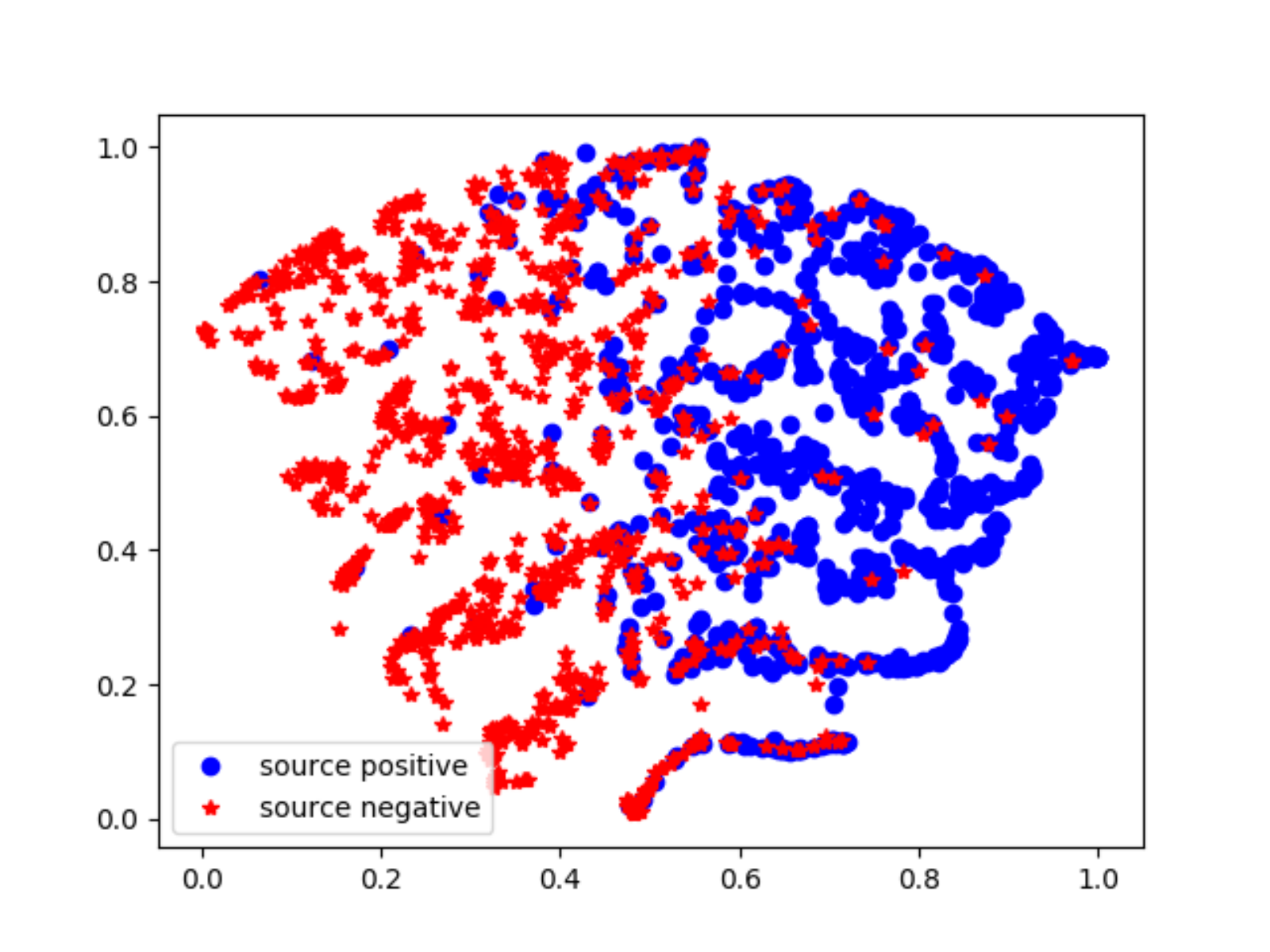}
\end{minipage}
}%
\subfigure[Target domain in HAGAN.]{
\begin{minipage}{0.45\linewidth}
\centering
\includegraphics[width=240px]{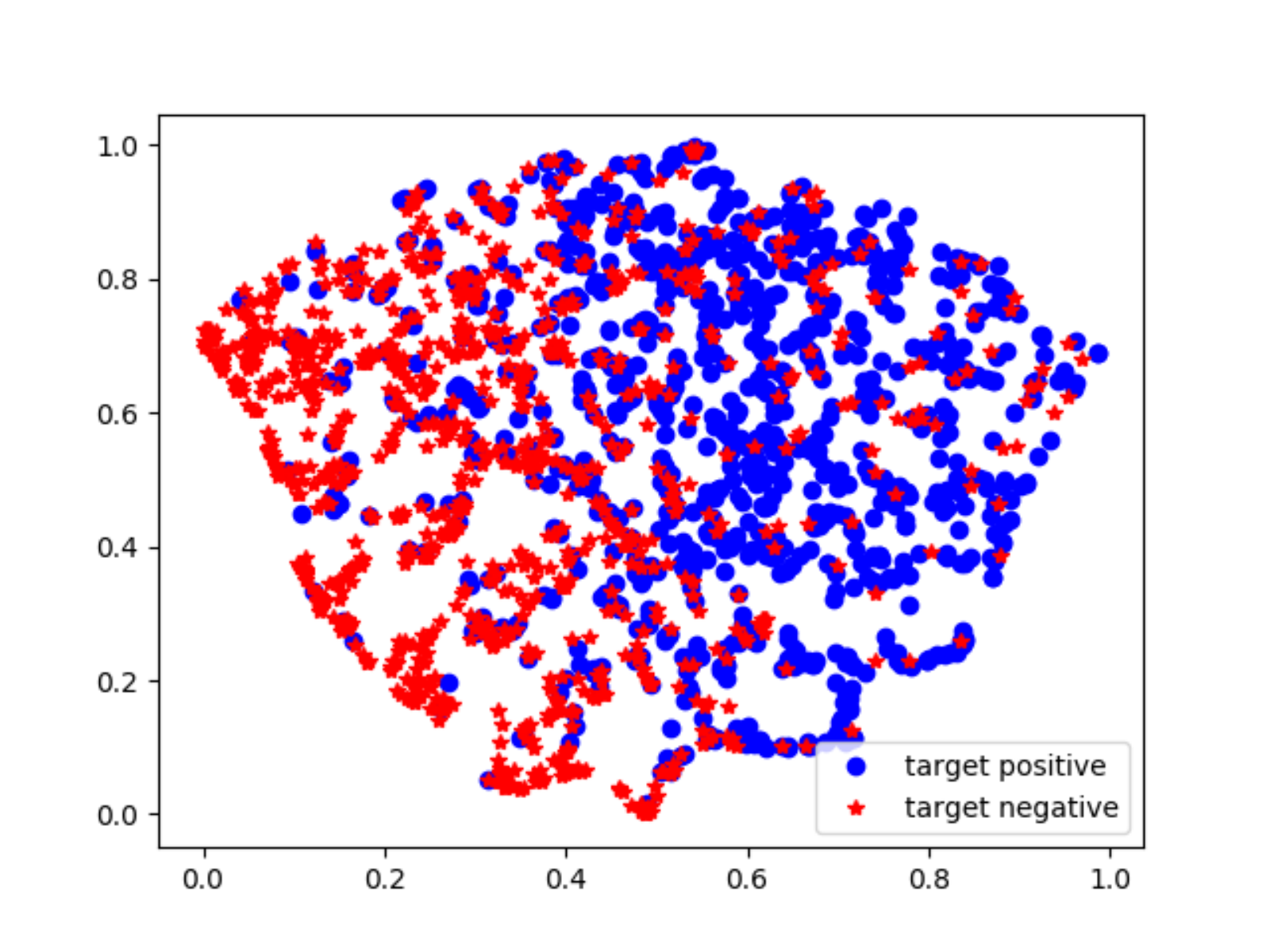}
\end{minipage}
}%

\centering
\caption{Visualization of document representations in B$\to$E task, which is implemented by t-SNE.}\label{fig2}
\end{figure*}

We show the visualization of the document representations in HAGAN in this section. As shown in Fig.~\ref{fig2} (a), the representation distributions of source and target domain data are very different in the naive HAN model which is trained on only source domain data (i.e. $\lambda_D = \lambda_G^1 = \lambda_G^2 = 0$). It means the sentiment classifier trained on the source domain data may not work well on the target domain data. However, as shown in Fig.~\ref{fig2} (b), the representation distributions of source and target domain data are almost identical, which means the sentiment classifier in the HAGAN model is domain-shared. The source and target domain data representation distributions are shown in Fig.~\ref{fig2} (c) and (d), respectively. We see that the HAGAN model enables sentiment identification of the representation distributions in both source and target domains.

\subsection{Visualization of Attention}

\begin{figure*}
  \centering
  \includegraphics[width=450px]{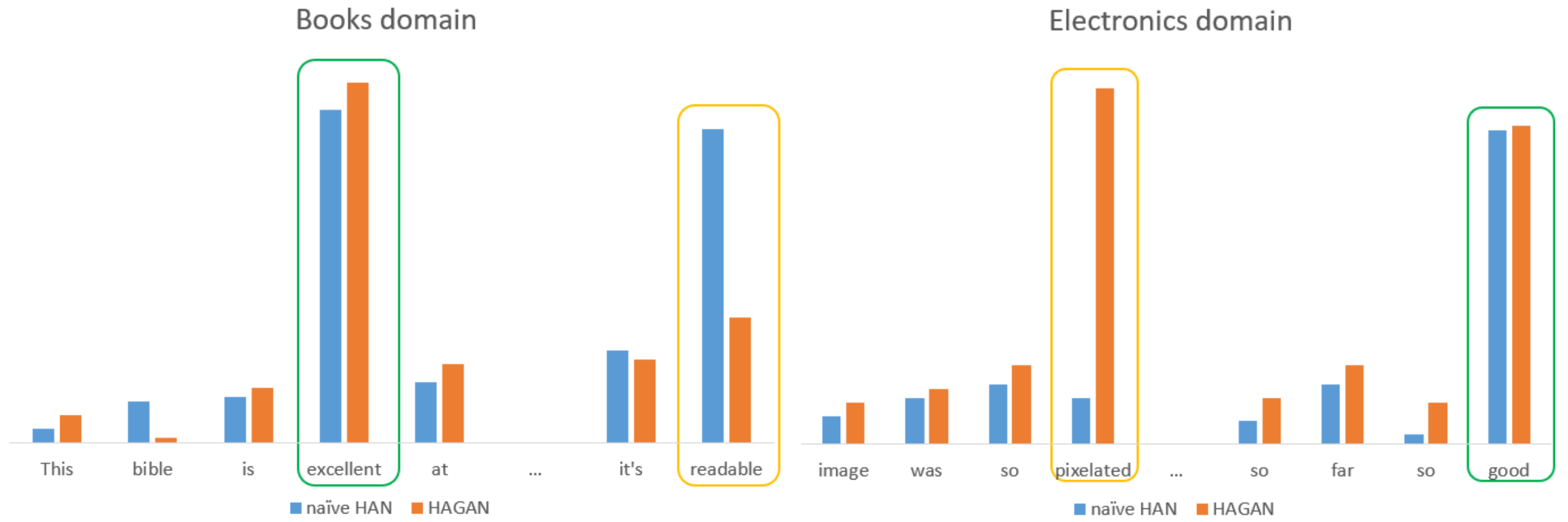}\\
  \caption{Visualization of attention in B$\to$E task.}\label{fig3}
\end{figure*}

In this section, we show the visualization of attention and describe how to capture the pivots and non-pivots based on the attention values in the HAGAN model. We give an example in Fig.~\ref{fig3} where the attention value is calculated by multiplying the current word attention value by the current sentence attention value. The naive HAN is a degenerate HAGAN where $\lambda_D = \lambda_G^1 = \lambda_G^2 = 0$, which means only the source domain labeled data is used for training. We determine a word as pivot if both the naive HAN and HAGAN give high attention values, e.g., \emph{excellent} and \emph{good} in Fig.~\ref{fig3}. We determine a word as non-pivot in source domain if the naive HAN gives high attention value while the HAGAN model gives low attention value, e.g., \emph{readable} in books domain. We determine a word as non-pivot in target domain if the naive HAN gives low attention value while the HAGAN model gives high attention value, e.g., \emph{pixelated} in electronics domain. Table~\ref{tab2} and Table~\ref{tab3} list some examples of pivots and non-pivots in B$\to$E task, which are captured based on the above rules.
\begin{table}
  \centering
  \begin{tabular}{c|c}
     \hline
     Positive & Negative \\
     \hline
     great good excellent  & bad poor disappointed   \\
     best beautiful amazing  & boring annoying  tedious  \\
     enjoyable love funny  & waste slow ridiculous \\
     nice inspiring humorous  & lacking shallow flawed \\
     true well interesting & misleading worst horrible \\
     perfect pretty real & difficult horrible dull \\
     \hline
  \end{tabular}
  \caption{Samples of pivots captured by the HAGAN model in B$\to$E task.}\label{tab2}
\end{table}
\begin{table}
  \centering
  \begin{tabular}{c|c|c}
     \hline
     Domain & Positive & Negative \\
     \hline
     \multirow{5}{*}{B}& readable believable  & unappealing insulting  \\
     & memorable genuinely  & trite disorganized  \\
     & eloquent  endearing & repetitious pointless \\
     & hearted understandable   & devoid sophomoric  \\
     & thoughtful appealing & forgettable distracting \\
     \hline
     \multirow{5}{*}{E}& useable noticeable & spotty scratched   \\
     & rubbery stereo  & plugged laborious  \\
     & pixelated illuminated  & negligible kludgy \\
     & prerecorded audible  & blurry oily \\
     & coaxed craving & noisy inferior \\
     \hline
  \end{tabular}
  \caption{Samples of non-pivots captured by the HAGAN model in B$\to$E task.}\label{tab3}
\end{table}
\section{Conclusion}
In this paper, we proposed the HAGAN model for cross-domain sentiment classification. The proposed HAGAN model applies GAN architecture instead of the GRL to adversarially train a generator and a discriminator, where the discriminator has two objectives which are sentiment classification and domain classification, and the objective of the generator is to generate a representation which is sentiment-distinguishable and domain-indistinguishable. The generator consists of two layers of Bi-GRU with hierarchical attention mechanism which map to word-sentence and sentence-document, respectively. The Bi-GRU help to encode the context information into the representation, and the attention mechanism help to capture the pivot and non-pivot automatically. The experiments on the Amazon review dataset show the effectiveness of the HAGAN model. The proposed HAGAN model could be potentially adapted to other domain adaption tasks, which is the focus of our future studies.
\section*{Acknowledgments}
The work described in this paper has been supported by the Ministry of Science and Technology (Grant No. 213), the Natural Science Foundation of China (Grant No. 61673301), the Major Project of Ministry of Public Security (Grant No. 20170004).
\bibliographystyle{named}
\bibliography{ijcai19}

\end{document}